\documentclass{article}
\usepackage{ijcai19}

\usepackage{times}
\usepackage{soul}
\usepackage{url}
\usepackage[hidelinks]{hyperref}
\usepackage[utf8]{inputenc}
\usepackage[small]{caption}
\usepackage{graphicx}
\usepackage{subcaption}
\usepackage{amsmath}
\usepackage{booktabs}
\usepackage{algorithm}
\usepackage{algorithmic}
\usepackage{natbib}
\usepackage{blkarray}
\usepackage{xfrac}
\usepackage{xcolor}

\DeclareMathOperator*{\argmax}{arg\,max}

\usepackage{mathtools}
\DeclarePairedDelimiter{\ceil}{\lceil}{\rceil}

\usepackage{scalerel,stackengine}
\stackMath
\newcommand\reallywidehat[1]{%
\savestack{\tmpbox}{\stretchto{%
  \scaleto{%
    \scalerel*[\widthof{\ensuremath{#1}}]{\kern-.6pt\bigwedge\kern-.6pt}%
    {\rule[-\textheight/2]{1ex}{\textheight}}
  }{\textheight}%
}{0.5ex}}%
\stackon[1pt]{#1}{\tmpbox}%
}

\title{Safe Coordination of Human-Robot Firefighting Teams}

\author{
Esmaeil Seraj
\and
Andrew Silva\and
Matthew Gombolay
\affiliations
Institute for Robotics and Intelligent Machines, Georgia Institute of Technology
\emails
eseraj3@gatech.edu,
\{andrew.silva, matthew.gombolay\}@cc.gatech.edu
}

\begin{document}

\maketitle

\begin{abstract}
Wildfires are destructive and inflict massive, irreversible harm to victims’ lives and natural resources. Researchers have proposed commissioning unmanned aerial vehicles (UAVs) to provide firefighters with real-time tracking information; yet, these UAVs are not able to reason about a fire's track, including current location, measurement, and uncertainty, as well as propagation. We propose a model-predictive, probabilistically safe distributed control algorithm for human-robot collaboration in wildfire fighting. The proposed algorithm overcomes the limitations of prior work by explicitly estimating the latent fire propagation dynamics to enable intelligent, time-extended coordination of the UAVs in support of on-the-ground human firefighters. We derive a novel, analytical bound that enables UAVs to distribute their resources and provides a probabilistic guarantee of the humans’ safety while preserving the UAVs’ ability to cover an entire fire.
\end{abstract}

\section{Introduction}
\label{sec:introduction}
\noindent
Costly and massively destructive, wildfires  inflict irreversible damage to both victims’ lives and natural resources. According to the Insurance Information Institute~\citep{III2018facts}, more than 122,000 wildfires burned more than 24 million acres between 2016 and 2018 in the United States, killing more than 12,000 people and causing approximately \$20 billion in damage.

Fighting wildfires is a dangerous task and requires accurate online information regarding firefront location, size, scale, shape, and propagation velocity~\citep{martinez2008computer,stipanivcev2010advanced,sujit2007cooperative}. Historically, researchers have sought to extract images from overhead satellite feeds to estimate fire location information~\citep{fujiwara2002forest,kudoh2003two,casbeer2006cooperative}. Unfortunately,, the resolution of satellite images is typically too low to do any more than simply detect the existence of a fire~\citep{casbeer2006cooperative}. Firefighters need frequent, high-quality images (i.e., information updates) to monitor fire propagation, choose the most effective fighting strategy,, and most importantly, stay safe. 


Recent advances in UAV technology have opened up the possibility of providing real-time, high-quality fire information to firefighters. However, the control of UAVs in this volatile setting provides particular challenges. Effective cooperation normally requires that each individual UAV has a comprehensive knowledge regarding the state of every other team member in order to coordinate the team’s collective actions. This assumption is not always appropriate due to communication constraints and complexities. Accordingly, researchers typically seek to develop distributed coordination algorithms that reduce the need for explicit communication~\citep{beard2006decentralized,mcintire2016iterated,nunes2015multi,choi2009consensus}.
\begin{figure}[t!]
	\centering
	\includegraphics[width=0.8\columnwidth]{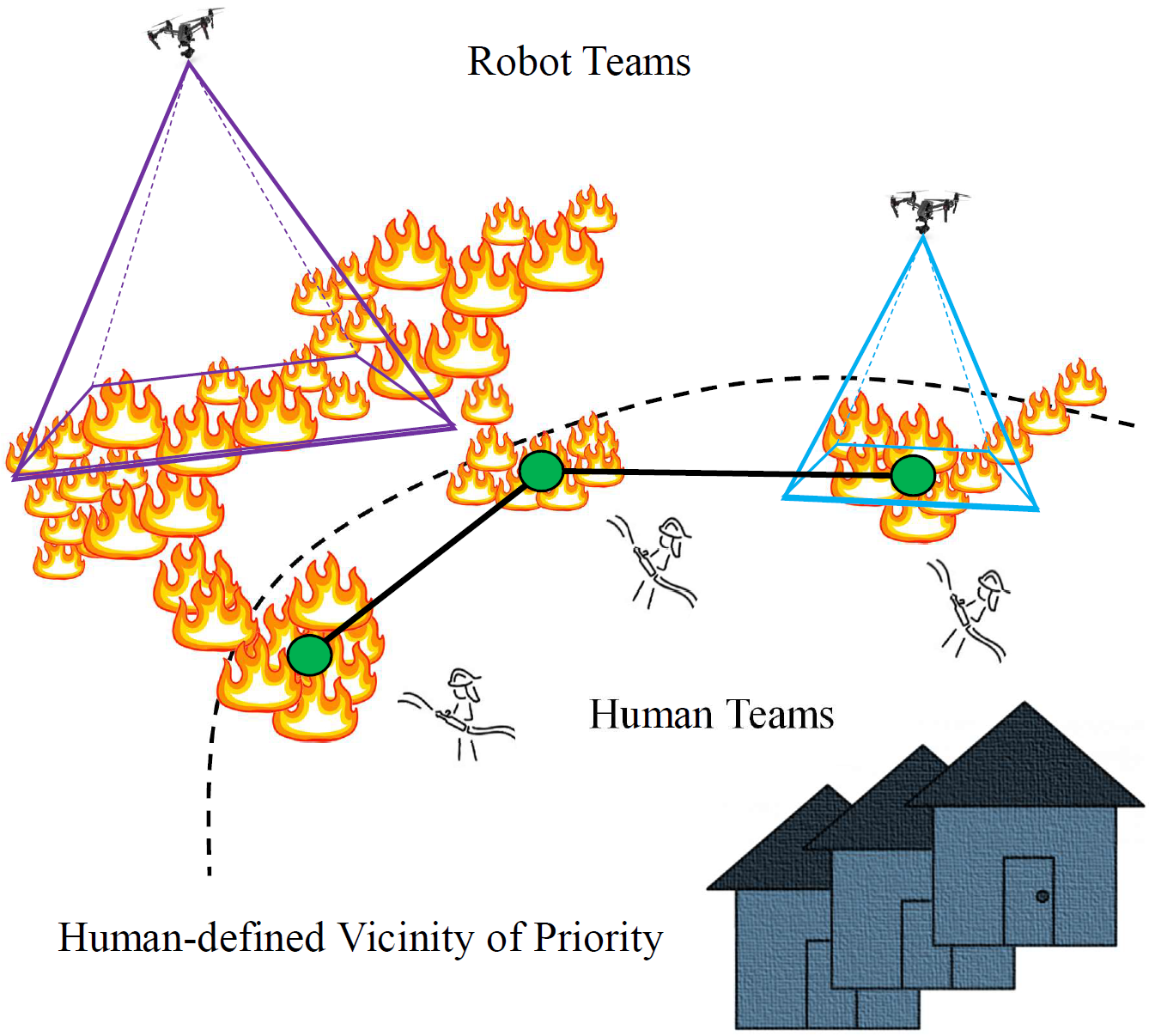}
	\caption{This figure depicts the proposed safe coordination of human-robot firefighting teams.}
	\label{fig:SafeHRI}
\end{figure}

Previous approaches to distributed control of UAVs have typically sought to maximize the ``coverage'' of a fire. In the works of \citet{schwager2011eyes}, and more recently \citet{pham2017distributed}, coverage is the average pixel density across an entire fire. However, these approaches do not explicitly reason (i.e., through tracking and filtering) regarding a fire state (i.e., position, velocity, and associated uncertainty), nor do they attempt to develop a predictive model for fire propagation by key parameters (e.g., fire-spread rate due to available fuel and vegetation). Furthermore, the location of a firefront is typically modelled as the ``coolest'' part of a fire visible via infrared sensors, but this is not always accurate~\cite{pastor2003mathematical}. 

In this paper, we overcome key limitations in prior work by developing an algorithmic framework to provide a model-predictive mechanism that enables firefighters on the ground to have probabilistic guarantees regarding their time-varying proximity to a fire. 
In our approach, we explicitly estimate the latent fire propagation dynamics via a multi-step adaptive extended Kalman filter (AEKF) predictor and the simplified FARSITE wildfire propagation mathematical model~\citep{finney1998farsite}. This model enables us to develop straightforward distributed control adapted from vehicle routing literature~\citep{cho2019traveling,laporte1992vehicle,toth2002vehicle,golden1980approximate} to enable track-based fire coverage.

In addition to this coverage controller, we derive a set of novel analytical bounds that allow UAVs to enable probabilistic-safe coordination of themselves in support of on-the-ground human firefighters. We investigate three different scenarios in which: (1) a wildfire is almost stationary, (2) a wildfire moves but does not grow (i.e., spawn) considerably, and (3) a wildfire moves quickly and multiplies. We derive analytic safe-to-work, probabilistic, temporal upper-bounds for each case to facilitate the firefighters’ understanding of how long they have until they must evacuate a location. We empirically evaluate our approach against simulated wildfires alongside contemporary approaches for UAV coverage~\citep{pham2017distributed}, as well as against reinforcement learning techniques, demonstrating a promising utility of our approach. 

\vspace{-0.15cm}
\section{Preliminaries}
\label{sec:preliminaries}
In this Section, we first introduce the simplified FARSITE wildfire propagation mathematical model and calculate the fire dynamics. We then review the fundamentals of AEKF.

\subsection{FARSITE: Wildfire Propagation Dynamics}
\label{subsec:simplifiedfarsite}
The Fire Area Simulator (FARSITE) wildfire propagation mathematical model was first introduced by \cite{finney1998farsite} and is now widely used by the USDI National Park Service, USDA Forest Service, and other federal and state land management agencies. The model has been employed to simulate a wildfire's spread, accounting for heterogeneous conditions of terrain, fuels, and weather and their influence on fire dynamics. The full FARSITE model includes complex equations for firefront location updates, and precise model implementation requires a significant amount of geographical, topographical, and physical information on terrain, fuels, and weather. Moreover, the precise implementation of the FARSITE model is computationally expensive, and thus, researchers tend to modify the model by considering several simplifying assumptions~\cite{pham2017distributed}. The wildfire propagation dynamics using a simplified FARSITE model is shown in Equations \ref{eq:firedynamics1} and \ref{eq:qdotdefinition}.
\par\nobreak{\parskip0pt \footnotesize \noindent
	\begin{align}
		q_t^i &= q_{t-1}^i + \dot{q}_{t-1}^i\delta t \label{eq:firedynamics1}\\
		\dot{q}_{t}^i &= \frac{d}{dt}\left(q_t^i\right)\label{eq:qdotdefinition}
\end{align}}

In the above Equations, $ q_t^i $ indicates the location of firefront $ i $ at time $ t $. $ \dot{q}_{t}^i $ identifies a fire’s growth rate (i.e., fire propagation velocity) and is a function of fire spread rate ($ R_t $, e.g., fuel and vegetation coefficient), wind speed ($ U_t $), and wind azimuth ($ \theta_t $), which are available to our system through weather forecasting equipment at each time $ t $. 

\subsection{Adaptive Extended Kalman Filter (AEKF)}
\label{subsec:EKF}
The Kalman filter is an optimal estimation method for linear systems with additive independent white noise in both transition and observation systems~\citep{kalman1960contributions,kalman1961new,kalman1960new}. For nonlinear systems, an extended Kalman filter (EKF), which adapts multivariate Taylor series expansion techniques to locally linearize functions, can be used. In an EKF, the state transition and observation models are not required to be linear but may instead be functions that are differentiable and therefore linearizable, as shown in Equations \ref{eq:syseq1} and \ref{eq:syseq2}, where $ f(s_{t-1}, u_t) $ and $ h(s_t) $ are the state transition and observation models, respectively connecting states $ s_t $ and $ s_{t-1} $ at $ t-1 $ and $ t $. 
\vspace{-.1cm}

\par\nobreak{\parskip0pt \footnotesize \noindent
	\begin{align}
		s_t &= f(s_{t-1}, u_t) + \omega_t \label{eq:syseq1} \\
		z_t &= h(s_t) + \nu_t \label{eq:syseq2}
\end{align}}
\vspace{-.35cm}

Taking $ u_t $ as control input, function $ f $ computes the predicted state at time $ t $ from the previous estimate at time $ t-1 $, where $ h $ maps the predicted state estimate to predicted state observation. Variables $ \omega_t $ and $ \nu_t $ model the process and observation noises, respectively, and are assumed to be zero mean multivariate Gaussian noise with covariances $ Q_t $ and $ R_t $.

\subsubsection{AEKF: Prediction Step}
\label{subsubsec:ekfpredictionstep}
The first step in EKF is the prediction step during which a prediction of the next to-be-observed state is estimated. The predicted state estimate is measured using Equation \ref{eq:ekfstateestimate}, and the predicted covariance estimate is calculated, as shown in Equation \ref{eq:ekfcovarianceestimate}, where $ F_t $ is the state transition matrix and defined in Equation \ref{eq:statetransitionjacobian} as the Jacobian of the fire propagation model (Equation \ref{eq:statetransitionjacobian}).
\par\nobreak{\parskip0pt \footnotesize \noindent
	\begin{align}
		\hat{s}_{t|t-1} &= f(\hat{s}_{t-1|t-1}, u_t) \label{eq:ekfstateestimate}\\
		P_{t|t-1} &= F_tP_{t-1|t-1}F_t^T + Q_t \label{eq:ekfcovarianceestimate} \\
		F_t &= \left. \frac{\partial f}{\partial s}\right|_{\hat{s}_{t-1|t-1}, u_t} \label{eq:statetransitionjacobian} \\
		\hat{s}_{t+r|t} &= F_t^{r-1}s_{t|t-1}\label{eq:multistepstatepred}
\end{align}}It can be shown that a multi-step prediction of the state from \textit{r} steps ahead $ \hat{s}_{t+r|t} $ can be obtained given the measurements and observations at time $ t $~\citep{taragna2011kalman}, as shown in Equation \ref{eq:multistepstatepred}.

\subsubsection{AEKF: Update Step}
\label{subsubsec:ekfupdatestep}
The state and covariance estimates are updated in this step, noted in Equations \ref{eq:ekfstateestimateupdate}-\ref{eq:ekfcovarianceestimateupdate}, where $ K_t = P_{t|t-1}H_t^TS_t^{-1} $ is the near-optimal Kalman gain and $ \tilde{y}_t = z_t - h(\hat{s}_{t|t-1}) $ is the measurement residual. Moreover, $ S_t $ is the covariance residual and is measured in Equation \ref{eq:residualkalman}, where $ H_t $ is the observation Jacobian matrix calculated  in Equation \ref{eq:observationjacobian}. Similar to Equation~\ref{eq:multistepstatepred}, the measurement residual of the observation model $ h(s_t) $ at  \textit{r} steps into the future is obtained using Equation \ref{eq:multistepresidual}.
\begin{align}
	\hat{s}_{t|t} &= \hat{s}_{t|t-1} + K_t\tilde{y}_t \label{eq:ekfstateestimateupdate} \\
	P_{t|t} &= \left(I - K_tH_t\right)P_{t|t-1} \label{eq:ekfcovarianceestimateupdate}\\
	S_t &= H_t P_{t|t-1}H_t^T + R_t \label{eq:residualkalman} \\
	H_t &= \left. \frac{\partial h}{\partial s}\right|_{\hat{s}_{t|t-1}} \label{eq:observationjacobian} \\
	S_{t+r|t} &= H_tF_t^{r-1}P_{t|t-1}\left(H_tF_t^{r-1}\right)^T + R_t \label{eq:multistepresidual}
\end{align}

We leverage AEKF~\citep{akhlaghi2017adaptive}, which introduces innovation and residual-based updates for process and observation noise covariances, as shown in Equations \ref{eq:adaptiveQ}-\ref{eq:adaptiver}. These updates remove the assumption of constant covariances $Q_t$ and $R_t$.
\begin{align}
	Q_t &= \alpha Q_{t-1} + (1-\alpha)\left(K_td_td_t^TK_t^T\right) \label{eq:adaptiveQ} \\
	R_t &= \alpha R_{t-1} + (1-\alpha)\left(\tilde{y}_t\tilde{y}_t^T + H_tP_{t|t-1}H_t^T\right) \label{eq:adaptiver}
\end{align}
$ Q_t $ and $ R_t $ are adaptively estimated as the Kalman filter leverages its observations to improve the predicted covariance matrix $ P_{t|t-1} $ by applying Equations \ref{eq:adaptiveQ}-\ref{eq:adaptiver}. This AEKF's multi-step prediction and error propagation provide the capability to derive an analytical condition for the safety of human firefighters, as discussed in Section \ref{subsec:humansafetymodule}. To the authors' knowledge, this has not yet been explored in the related literature.


\vspace{-0.15cm}
\section{Method}
\label{sec:method}
The proposed algorithm for safe human-robot coordination in wildfire fighting is henceforth summarized. Initially, all available drone resources are distributed to cover a wildfire by generating a search graph and clustering the hotspots to generate an optimal path for each drone. Hotspots are detected through vision or thermal cameras~\citep{merino2010automatic,yuan2015survey}. We assume that firefighters' locations are known to the system through common locating devices, such as GPS. Upon receiving a support request from a human team, one of the drones conducts a ``feasibility test'' based on an analytical, probabilistic bound we derived in Section \ref{subsec:T_UB}, Equations \ref{eq:T_UBcase1}, \ref{eq:T_UBcase2}, and \ref{eq:T_UBCase3} to determine whether the UAV can safely protect the human firefighters. This bound determines an upper bound on the time, $T_{UB}$, required to cover each fire in the vicinity of the human firefighters. $T_{UB}$ is compared to the maximum time allowable for each fire track to propagate before measurement—derived from the AEKF model—so that the residual post-measurement is less than the residual after the last time a given firefront location was observed. This test is presented in Section \ref{subsubsec:URR} with Equation \ref{eq:URR1}. If the test is satisfied, a near-optimal tour is computed via a k-opt procedure. If the test fails, the UAV recruits more UAVs to assist until this bound is satisfied, divvying up the responsibilities for covering the fire locations. Next, we describe in detail each component of our overall method.



The proposed framework depends on the calculated upper bound traverse time, $ T_{UB} $, which itself is dependent on the propagation velocity and growth rate of the wildfire. Details of different wildfire scenarios and the analytical traverse time upper bound for each case are presented in the following section.

\subsection{Probabilistic Safety for Coordination}
\label{subsec:T_UB}
\begin{figure}[t!]
	\centering
	\includegraphics[trim=0in 0in 0in 0in,clip,width=\columnwidth]{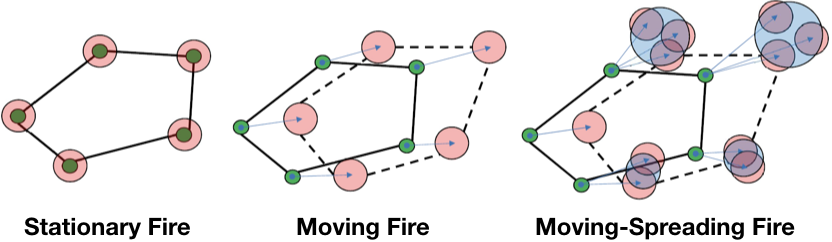}
	\caption{This figure depicts the three wildfire scenarios we consider. case 1: near-stationary (Left), case 2: wildfire moves but does not grow considerably (Center), and case 3: wildfire moves quickly and multiplies (Right). The green dots and red circles represent current and next-step locations of firefronts, and the blue circles indicate the Steiner zones.}
	\label{fig:WildfireScenarios}
\end{figure}

In this section, we develop a probabilistic upper bound, $T_{UB}$, on the time required by a drone to service each fire location (i.e., observe once). This upper bound is then used in Section \ref{subsubsec:URR}, Equation \ref{eq:URR1} to determine whether a drone can be probabilistically guaranteed to service each fire location fast enough to ensure a bounded track residual for each fire. We derive three bounds, one for each of the following fire scenarios: (1) a wildfire is almost stationary, (2) a wildfire moves but does not grow (i.e., spawn) considerably, and (3) a wildfire moves quickly and multiplies. Figure~\ref{fig:WildfireScenarios} depicts these scenarios. 

For the derivation, we reason about the velocity of the fire at the $\alpha$ confidence level (i.e., the probability the velocity is greater is $1-\alpha$). We set $\alpha=0.05$ for our experiments. Further, we assume each fire's velocity is conditionally independent given latent fire dynamics, which we estimate explicitly in our AEKF model. As such, the probability that our bounds are correct is demonstrated in Equation \ref{eq:P_Correct}, where $T^*$ is the actual maximum time the drone can go without visiting any $q \in Q$ and $T_{UB}$ is our bound. \footnote{A formal proof and derivations is included in our supplementary
	version, available at https://github.com/firefront-tracking/Safe-
	Coordination-of-Human-Robot-Firefighting-Teams}
\par\nobreak{\parskip0pt \footnotesize \noindent
	\begin{equation}
		Pr[{T_{UB} \geq T^*}] \leq 1- \prod_{q \in Q} 1-\alpha \label{eq:P_Correct}
\end{equation}}

\subsubsection{Case 1: Stationary Fire}
\label{subsubsec:stationary-fire}
The first scenario is that of a fire which is almost stationary. Considering a search graph made of all hotspots to visit as vertices and paths to traverse as edges, we calculate an initial path using a minimum spanning tree (MST). Note that the MST path (i.e., the Hamiltonian Cycle computed from the MST) is not the fastest possible path, but it is fast to generate. Therefore, the drones utilize the k-opt algorithm to iteratively improve the path and bring it closer to optimal. This process ensures a quick, sound way to guarantee the safety for the people on the ground. Once a consensus is reached with respect to the safety of human firefighters, UAVs can spend time improving the efficiency of actual firefront tours. We derive the upper bound time to complete a tour to all locations as the Hamiltonian path of the MST path (denoted with temporal cost $MST$) for a drone with velocity $ v$ in Equation \ref{eq:T_UBcase1}.
\par\nobreak{\parskip0pt \footnotesize \noindent
	\begin{equation}
		\label{eq:T_UBcase1}
		T_{UB}^{(\text{C1})} = \frac{2MST}{v}
\end{equation}}
\vspace{-0.15cm}

\subsubsection{Case 2: Moving Fire}
\label{subsubsec:slow-moving-fire}
The second scenario for consideration is a fire that moves but does not grow (i.e., spawn) considerably. In this case, the number of nodes $ |Q| $ (i.e., discretized firefront points) and the velocity of the propagating fire need to be taken into account. Our upper bound on the time a UAV has to revisit each fire location is noted in Equation \ref{eq:T_UBcase2}, where {\tiny{$\left.\reallywidehat{\dot{q}_t^\xi}\right|_{\alpha}$}} is the estimated speed of fire $q$ in the $\xi\in{x,y}$ direction evaluated at confidence interval defined by $\alpha$. 
\par\nobreak{\parskip0pt \footnotesize \noindent
	\begin{align}
		T_{UB}^{(\text{C2})} &= \frac{{MST}}{\frac{v_m}{2} - 2\zeta^{\alpha}\left(|Q| - 1\right)} \label{eq:T_UBcase2} \\
		\zeta^{\alpha} &= \argmax_{q,q'} \sqrt{\left(\left.\reallywidehat{\dot{q}_t^x}\right|_{\alpha}\right)^2 + \left(\left.\reallywidehat{\dot{q'}_t^y}\right|_{\alpha}\right)^2}\label{eq:U95}
\end{align}}
In Equation \ref{eq:T_UBcase2}, we assume a worst-case speed for all fires propagating at a rate of the speed of the  fastest fire in the x- and y-directions, as shown in Equation \ref{eq:U95}. By setting $\alpha$, we can control the degree of confidence in our system at the cost of making the UAV coordination problem more difficult. This independence assumption and uniformity of fastest and universal speed is a worst-case assumption. 

To arrive at this bound, we begin by considering the cost of traveling to each fire under the stationary case, $T_{UB}^{(C1)}$. If the fire locations are moving, the edges of the Hamiltonian cycle connected to each fire location will shrink or expand. In the worst case, we assume that each edge expands according to two times the velocity of the fastest fire location. Considering that the drone has $|Q|$ fires to consider, this yields $(|Q|-1)$ edges that may expand in the MST, with two times that quantity accounting for the Hamiltonian path. We note that the total expansion is a function of the time the fire is able to expand, which is a self-referencing equation, as shown in Equation \ref{eq:T_UBcase2_start}. However, we easily solve for $T_{UB}^{(\text{C2})}$ to arrive at Equation \ref{eq:T_UBcase2}.
\par\nobreak{\parskip0pt \footnotesize \noindent
	\begin{align}
		T_{UB}^{(\text{C2})} = T_{UB}^{(\text{C1})} + 2*\left(|Q| - 1\right)\zeta^{\alpha}*T_{UB}^{(\text{C2})} \label{eq:T_UBcase2_start}
\end{align}}

\vspace{-.15cm}

\subsubsection{Case 3: Moving-Spreading Fire}
\label{subsubsec:rapidly-growing-fire}
The third scenario we consider is a wildfire that moves and grows quickly. Single nodes of fire expand over time and escape the drone's field of view (FOV). The drones must consider the time it takes for a spawning point $ q $ to escape their FOV, given its velocity and the width of the FOV $ g $. The upper bound for traversal time allowed to maintain the track quality of each fire is shown in Equation \ref{eq:T_UBCase3}, where $\gamma =  \frac{4|Q|\left(\zeta^{\alpha}\right)^2}{gv}$, $\beta = 1 + \frac{2|Q|\zeta^{\alpha}}{v}$ and $\delta = \frac{MST}{\frac{v}{2} - 2\zeta^{\alpha}(|Q|-1)}$.
\begin{align}
	T_{UB}^{(C3)} &= \frac{-\beta + \sqrt{\beta^2 - 4\gamma\delta}}{2\gamma} \label{eq:T_UBCase3}
\end{align}
The width of the FOV for a drone $ d $ with altitude $ p_d^z $ is the half-angle $ \theta $ of the camera as $g = 2p_d^z\tan\theta$.

To arrive at this bound, we begin by considering the time to traverse to each moving firefront and the time to cover the area possibly engulfed by the growth of each firefront, as shown in Equation \ref{eq:T_UBCase3_start}. The first term of this equation brings $T_{UB}^{(\text{C2})}$ from Equation \ref{eq:T_UBcase2}. However, we must also account for the fact that the fires are able to spread for $T_{UB}^{(\text{C3})}$ units of time instead of only $T_{UB}^{(\text{C2})}$. Second, we must consider how much growth a fire location will experience. For a specific fire, $q$, the fire will grow no more than {\tiny{$2\left.\reallywidehat{\dot{q}_t^x}\right|_{\alpha}T_{UB}^{(C3)}$}} along the x-direction and no more than {\tiny{$2\left.\reallywidehat{\dot{q}_t^y}\right|_{\alpha}T_{UB}^{(C3)}$}} along the y-direction at the $\alpha$ confidence level. If we assume a vertical scanning pattern, the total number of passes the drone takes from left to right is given by {\tiny{$\ceil{ \sfrac{2\left.\reallywidehat{\dot{q}_t^x}\right|_{\alpha}T_{UB}^{(C3)}}{g}}$}}, and the total distance traversed for each pass is  {\tiny{$2\left.\reallywidehat{\dot{q}_t^y}\right|_{\alpha}T_{UB}^{(C3)}$}}.
\par\nobreak{\parskip0pt \footnotesize \noindent
	\begin{align}
		T_{UB}^{(C3)} &=  \left.T_{UB}^{(\text{C2})}\right|_{C3} +  \sum_q \frac{2}{v}\left.\reallywidehat{\dot{q}_t^y}\right|_{\alpha}T_{UB} \ceil[\Bigg]{ \frac{2\left.\reallywidehat{\dot{q}_t^x}\right|_{\alpha}T_{UB}^{(C3)}}{g}} \label{eq:T_UBCase3_start}
\end{align}}We make the following two conservative and simplifying assumptions. First, we remove the ceiling operator and add one to the term inside the operator to achieve continuity.
Second, we adopt a similar approach to case 2 by assuming the area required to cover to account for the growth of each fire location is upper-bounded by $|Q|$ times the area of growth for a hypothetical fire growing quickest.
With these two conservative simplifying assumptions, we readily arrive at the bound in Equation \ref{eq:T_UBCase3_start}.\footnote{See Appendix}

\subsection{Human Safety Module}
\label{subsec:humansafetymodule}
\subsubsection{Close-enough Traveling Salesman Problem}
\label{subsubsec:CETSP}
The first step in our human safety module is to generate a search graph with firefront points within human vicinity as the vertices and distances in between as the edges. For this purpose, we leverage the close-enough traveling salesman problem (CE-TSP) with Steiner zone~\cite{wang2019steiner} variable neighborhood search. In CE-TSP, the agent does not need to visit the exact location of each fire. Instead, the agent only gets ``close enough" to each goal.

In the application of monitoring firefront hotspots, a UAV observes multiple nodes within its FOV and thus, does not need to travel to the exact location of each firefront point. Instead, we first identify the overlapping disks between $ k $ fire points as their corresponding Steiner zone~\cite{wang2019steiner} and choose a node from the identified Steiner areas as a new vertex in our tour instead of the corresponding $ k $ points. 

\subsubsection{Analytical Safety Condition: Uncertainty Residual Ratio}
\label{subsubsec:URR}
Upon reaching a new node within the human vicinity, the UAV calculates two quantities: (1) the time required to complete a tour generated in the previous step using the analytical time upper bound Equations $T_{UB}$ and (2) the current measurement residual covariance through Equation~\ref{eq:residualkalman}. Leveraging these two parameters, now Equation~\ref{eq:multistepresidual} is used to predict an estimation of the residual $ T_{UB} $ steps into the future. We introduce the uncertainty residual ratio (URR) in Equation \ref{eq:URR1}.
\vspace{-0.3cm}

\par\nobreak{\parskip0pt \footnotesize \noindent
	\begin{equation}
		URR_t^q = \frac{S_{t+T_{UB}|t}}{S_{t|t-1}} \leq 1, \forall q \in Q \label{eq:URR1}
\end{equation}}

The $ URR_t^q $ is an indicator of the scale to which the UAV is capable of tracking the fire point $ q $ within the vicinity of the humans without losing tracking information. A URR greater than one demonstrates a quickly growing or propagating wildfire towards the human team, about which the UAV will not be capable of providing online information while completing a tour before the situation becomes dangerous and retreat is necessary. Similarly, a URR smaller than one indicates that online information can be provided while keeping track of the fire and ensuring the safety of humans.

If the URR condition is not satisfied for a node, the generated path in the CE-TSP step is partitioned into two paths and another UAV is recruited from the available resources. This process is repeated until $ URR_t^q $ is satisfied for all $ q $ and $ t $. 
To calculate the residuals as in the URR Equation in~\ref{eq:URR1}, we use the adaptive extended Kalman filter (AEKF). The process and observation model derivations of AEKF are detailed in the following sections.

\subsection{Uncertainty and Residual Prediction}
\label{subsec:errorpropagation}
For the purpose of measurement residual estimation, we use EKF to predict a distribution and accordingly, a measurement covariance for each firefront point through linear error propagation techniques. 
Considering $ q_{t-1} $ as the location of firefronts at current time and $ p_{t-1} $ as the UAV coordinates, a firefront location $ \hat{q}_t $ one step forward in time is desired, given the current firefront distribution ($ q_{t-1} $), fire propagation model with current parameters ($ \mathcal{M}_{t-1} $), and UAV observation model of the field ($ \mathcal{O}_{t-1} $). In other words, $\hat{q}_t = \argmax_{q_t} \rho\left(q_{t-1}, p_{t-1}, \mathcal{M}_{t-1}, \mathcal{O}_{t-1}, q_t\right)$.


We now compute the terms from the EKF equations presented in Section~\ref{subsec:EKF}, where wildfire propagation model $ \mathcal{M} $ and drone observation model $ \mathcal{O} $ are functions $ f $ and $ h $ in Equation~\ref{eq:syseq1}-\ref{eq:syseq2}. 
Using the aforementioned notations, the EKF state transition and observation equations can now be stated in Equations \ref{eq:probform1} and \ref{eq:probform3}, where $ p_{t-1} = [p_t^x, p_t^y, p_t^z] $ is the physical location of UAV. 
\par\nobreak{\parskip0pt \footnotesize \noindent
	\begin{align}
		q_t = & f_{t-1}\left(q_{t-1}, p_{t-1}, R_{t-1}, U_{t-1}, \theta_{t-1}\right) + \omega_t \label{eq:probform1} \\
		\hat{q}_t = & h_{t}\left(q_t, p_{t-1}\right) + \nu_t \label{eq:probform3}
\end{align}}We reform the state transition equation in~\ref{eq:probform1} to account for all associated parameters given by $ \mathcal{S}_t = \left[ q_t^x, q_t^y, p_t^x, p_t^y, p_t^z, R_t, U_t, \theta_t \right]^T $. Additionally, $ \Phi_t = \left[q_t^x, q_t^y, p_t^x, p_t^y, p_t^z, R_t, U_t, \theta_t\right]^T $ is a mapping vector through which the predicted fire dynamics and drone locations are translated into a unified angle-parameter vector {\footnotesize{$ \hat{\Phi}_t = \left[\varphi_t^x, \varphi_t^y, \hat{R}_t, \hat{U}_t, \hat{\theta}_t\right]^T $}}. The angle parameters (i.e., $ \varphi_t^x $ and $ \varphi_t^y $) contain information regarding both firefront location $ [q_t^x, q_t^y] $ and UAV coordinates $ [p_t^x, p_t^y] $. 
By projecting the looking vector of UAV to planar coordinates, the angle parameters are calculated as {\footnotesize{$\varphi_t^x = \tan^{-1}\left(\frac{q_t^x - p_t^x}{p_t^z}\right)$}} and {\footnotesize{$\varphi_t^y =  \tan^{-1}\left(\frac{q_t^y - p_t^y}{p_t^z}\right)$}}.
\begin{figure*}
	\centering
	\begin{subfigure}[t]{0.32\textwidth}
		\centering
		\includegraphics[width=\textwidth]{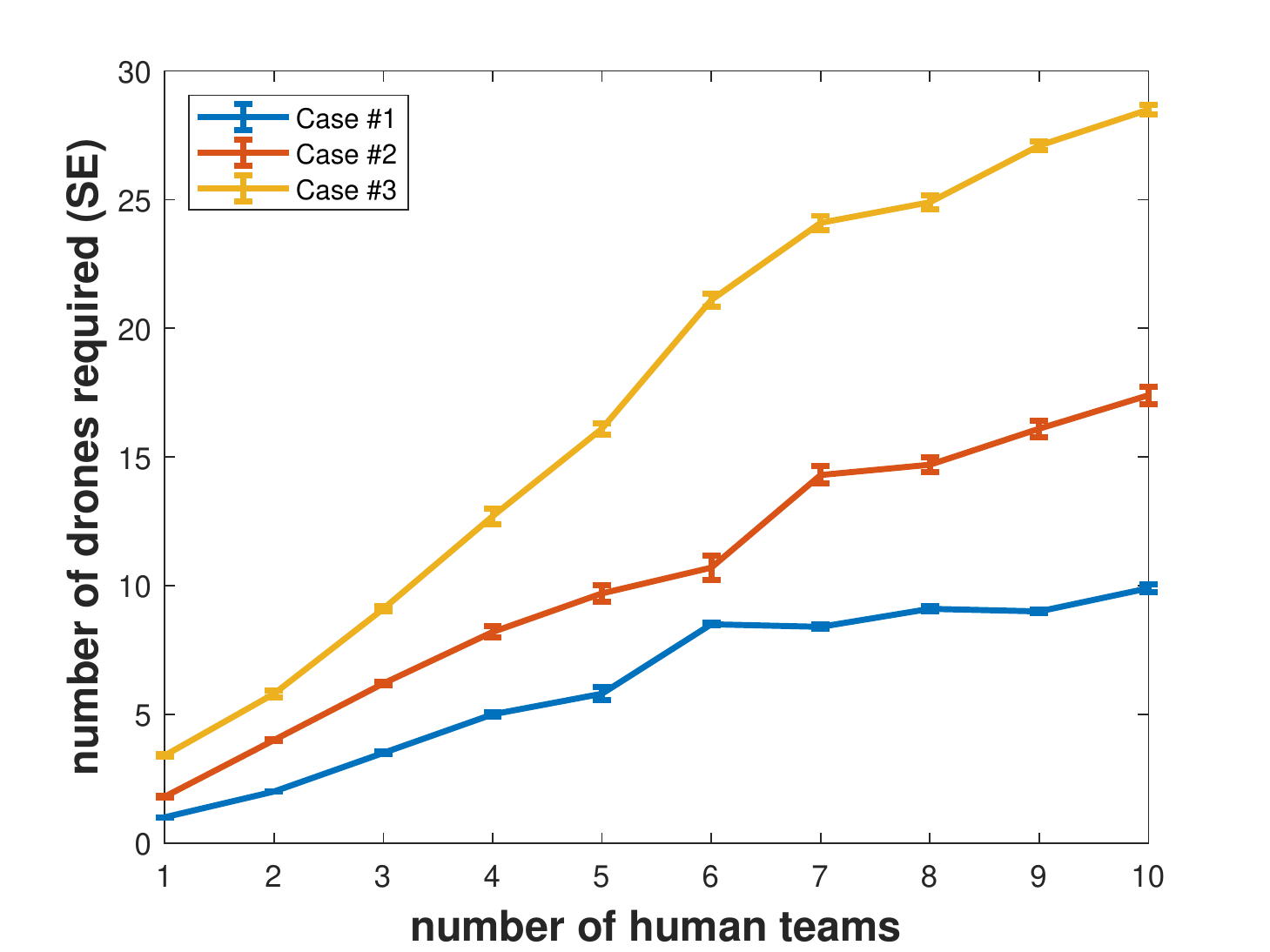}
		\caption{Drones required for human safety}
		\label{fig:num_drones_vs_num_human_teams}
	\end{subfigure}
	~~
	\begin{subfigure}[t]{0.32\textwidth}
		\centering
		\includegraphics[width=\textwidth]{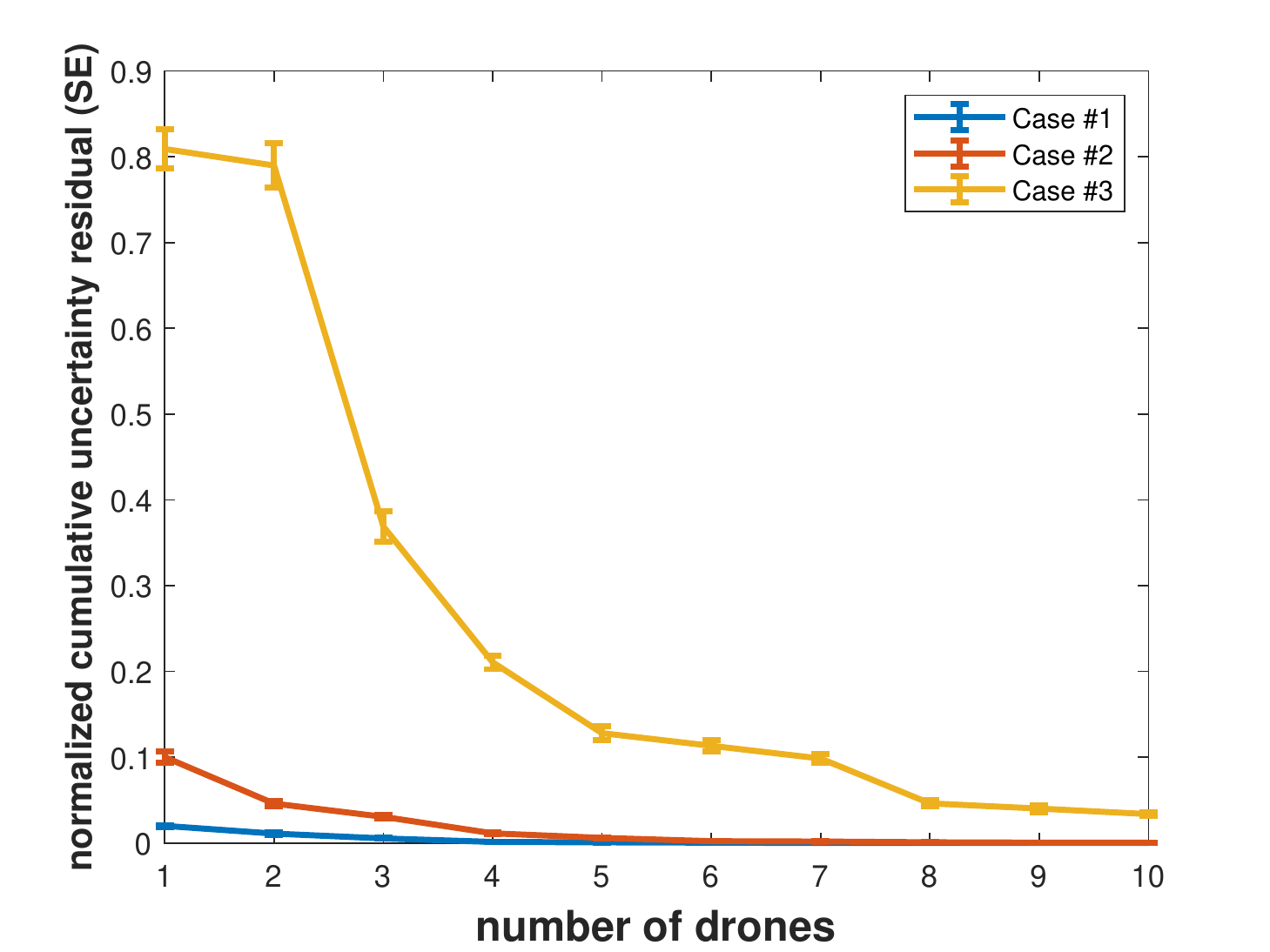}
		\caption{Fire-map uncertainty}
		\label{fig:num_drones_vs_uncertainty}
	\end{subfigure}
	~~
	\begin{subfigure}[t]{0.32\textwidth}
		\centering
		\includegraphics[width=\textwidth]{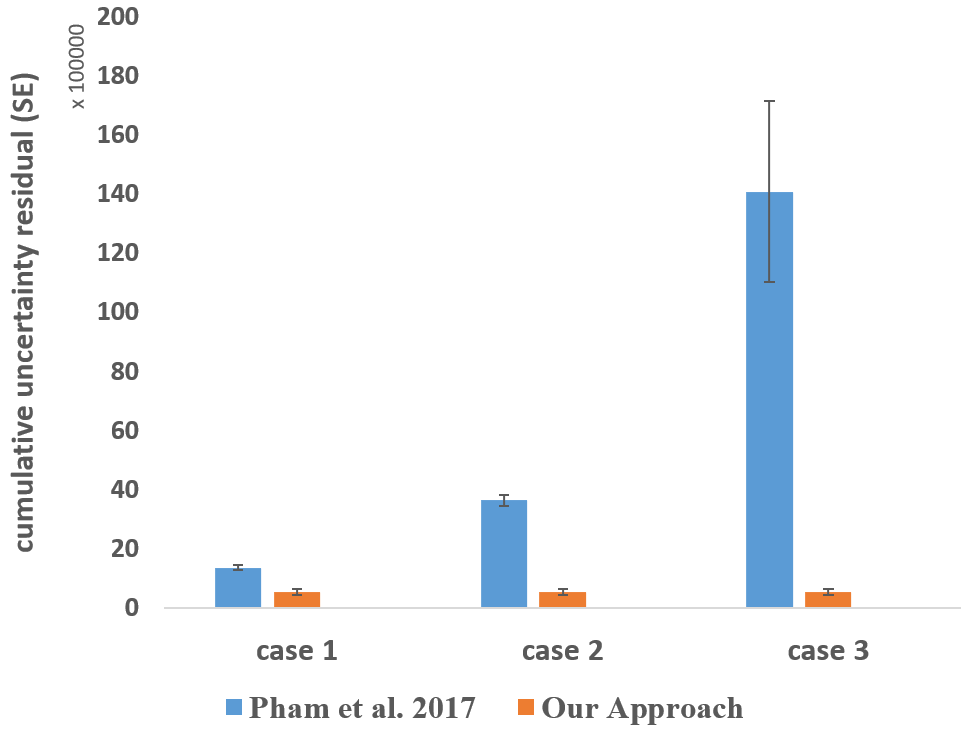}
		\caption{Comparison to prior work}
		\label{fig:comparison}
	\end{subfigure}
	\caption{This figure depicts a quantitative evaluation of the human safety analytical bound (Figure \ref{fig:num_drones_vs_num_human_teams}), the efficacy of the coverage coordination algorithm (Figure \ref{fig:num_drones_vs_uncertainty}), and a comparison to prior work (Figure \ref{fig:comparison}).}
\end{figure*}

The next step is to calculate the state transition Jacobian matrices $ F_t $ and $H_t$ in Equations \ref{eq:statetransitionjacobian} and \ref{eq:observationjacob}, respectively \footnote{See Appendix}, where $ t^\prime = t-1 $ and superscript $ (3) $ represent number of column and row repetitions for all $ [p_t^x, p_t^y, p_t^z] $.
\par\nobreak{\parskip0pt \scriptsize \noindent
	\begin{equation}
		\label{eq:statetransitionJacob}
		\left. \frac{\partial f}{\partial \mathcal{S}_i}\right|_{\hat{\Phi}_{t'}} =
		\begin{blockarray}{ccccccc}
			& q_{t^\prime}^x & q_{t^\prime}^y & p_{t^\prime}^{(3)} & R_{t^\prime} & U_{t^\prime} & \theta_{t^\prime} \\
			\begin{block}{c(cccccc)}
				q_t^x & 1 & 0 & 0^{(3)} & \frac{\partial q_{t}^x}{\partial R_{t^\prime}}  & \frac{\partial q_{t}^x}{\partial U_{t^\prime}} & \frac{\partial q_{t}^x}{\partial \theta_{t^\prime}} \\
				q_t^y & 0 & 1 & 0^{(3)} & \frac{\partial q_{t}^y}{\partial R_{t^\prime}}  & \frac{\partial q_{t}^y}{\partial U_{t^\prime}} & \frac{\partial q_{t}^y}{\partial \theta_{t^\prime}} \\
				p_{t}^{(3)} & 0 & 0 & 0^{(3)} & 0  & 0 & 0  \\
				R_t & 0 & 0 & 0^{(3)} & 1  & 0 & 0  \\
				U_t & 0 & 0 & 0^{(3)} & 0  & 1 & 0  \\
				\theta_t & 0 & 0 & 0^{(3)} & 0  & 0 & 1  \\
			\end{block}
		\end{blockarray}
\end{equation}}
\par\nobreak{\parskip0pt \tiny \noindent
	\begin{multline}
		\label{eq:observationjacob}
		\left. \frac{\partial h}{\partial \Phi_i}\right|_{\hat{\Phi}_{t|t-1}} = \\
		\begin{blockarray}{ccccccccc}
			& q_{t}^x & q_{t}^y & p_{t}^x & p_{t}^y & p_{t}^z & R_{t} & U_{t} & \theta_{t} \\
			\begin{block}{c(cccccccc)}
				\varphi_t^x & \frac{\partial \varphi_{t}^x}{\partial q_t^x} & 0 & \frac{\partial \varphi_{t}^x}{\partial p_t^x} & 0  & \frac{\partial \varphi_{t}^x}{\partial p_t^z} & 0 & 0 & 0 \\	
				\varphi_t^x & 0 & \frac{\partial \varphi_{t}^y}{\partial q_t^y} & 0 & \frac{\partial \varphi_{t}^y}{\partial p_t^y}  & \frac{\partial \varphi_{t}^y}{\partial p_t^y} & 0 & 0 & 0 \\
				\hat{R}_t & 0 & 0 & 0 & 0 & 0 & 1 & 0 & 0 \\
				\hat{U}_t & 0 & 0 & 0& 0 & 0 & 0 & 1 & 0 \\
				\hat{\theta}_t & 0 & 0 & 0 & 0 & 0 & 0 & 0 & 1 \\
			\end{block}
		\end{blockarray}
\end{multline}}

\subsection{Distributed Coverage}
\label{subsec:distributedcoverage}
While UAVs are not required to monitor the firefronts directly and provide human firefighters with safety information, they will explore the rest of the wildfire. As discussed earlier, distributed exploration and coverage is more efficient and practical as states of agents are hardly known to other agents due to communication constraints. In this section, we propose a distributed framework similar to our human safety module. The steps to our distributed coverage module are as follows: after detecting the fire map and hotspots using visual or thermal cameras, a set of fire points based on the aforementioned Steiner zone variable neighborhood search method are generated. The set of nodes are partitioned (e.g., K-means clustering) according to the number of available UAVs. Each drone is assigned to one partition by solving a simple constraint satisfaction problem (CSP) with drones as variables, partitions as domains, and distance to the center of the partition as constraints. Next, an optimal path for coverage and tracking is found applying the k-opt algorithm. The upper bound time, $ T_{UB} $, is then calculated for a fire point in the center of the drone's FOV. $T_{UB}$ provides the drone with an estimate of how long it will take afire to escape its FOV as determined by fire propagation velocity. When a route is identified in this way, UAVs can apply this reasoning for the next $ T_{UB} $ timesteps before recalculating a new path. After $T_{UB}$ has passed, the partitions are revised, and an optimal path is recalculated, since fire locations have more than likely changed significantly during this time. Once a drone is requested for human safety at firefronts, one of the drones is dismissed, and the process is repeated.

\section{Empirical Evaluation}
\label{sec:results}
\subsection{Benchmarks}
\label{subsec:benchmarks}
The first benchmark we compare to is the distributed control framework for dynamic wildfire tracking as proposed by \cite{pham2017distributed}. The distributed control framework includes two controller modules for field coverage and path planning in which the former is responsible for covering the entire fire and the latter for path planning. The two controllers are defined as the negative gradients (gradient descent) of objective functions to maximize the area-pixel density of the UAV's fire observations and to maintain safe flight parameters with potential field-based criteria, respectively. 



We also compare our approach to a reinforcement learning (RL) baseline. An RL policy network controls a single UAV, and each policy network is identical across UAVs. The network consists of four convolutional layers followed by three fully connected layers with ReLU activations. A single-channel image with fire locations is the input, and the network outputs a direction for each drone to move. The reward at each step is given as the negative sum of uncertainty across the entire map, encouraging the agent to minimize uncertainty over time. After more than 800 episodes, agents continue to exhibit random behavior and fail to find or track any firefronts. While we do not claim that RL cannot work in this application, we believe our learning approach is suited to the task yet is unsuccessful in producing meaningful results. As such, we do not include the RL baseline results in Figure \ref{fig:comparison}.

\subsection{Results}
\label{subsec:results}
In order to evaluate the efficacy of our proposed human safety module, we performed a comprehensive simulation to determine the number of drones needed to satisfy the uncertainty residual ratio (given in Equation~\ref{eq:URR1}) for a range of numbers of humans on a team. We performed the evaluation for all three aforementioned wildfire scenarios for 10 trials and calculated the mean and standard error (SE) for each. The results of these simulations are noted in Figure~\ref{fig:num_drones_vs_num_human_teams}. During our simulations, we chose the fire velocity $ \dot{q} $ to be 0, 0.5, and 1 for wildfire cases 1, 2, and 3, respectively. We also chose the spawning rate of the fire for case 3 to be at most 3, meaning each fire point can reproduce up to 3 more fire points. As shown in Figure~\ref{fig:num_drones_vs_num_human_teams}, increasing the number of operating human teams resulted in a rise in the number of drones required to guarantee safe conditions. This also held as the wildfire propagation scenario changed and a fire propagated more aggressively. 

We also evaluated our distributed coverage platform within a similar framework by calculating the cumulative uncertainty residual while covering the map. The uncertainty residual measure was an indicator of how successful UAVs were in covering all fire points and hotspots and increased by the number of nodes not covered by any drone. Figure~\ref{fig:num_drones_vs_uncertainty} shows the results for the evaluation of our distributed coverage method. For this evaluation, we required the drones to cover the entire fire map as much as possible, while maintaining an altitude that was both safe and conducive to high-quality imaging. As represented in Figure~\ref{fig:num_drones_vs_uncertainty}—the case of stationary wildfire—drones easily covered the fire map without missing too many fire points. In the case of a fast-growing fire, however, more drones were required to cover the fire map efficiently, reducing its associated entropy.

Figure~\ref{fig:comparison} shows a comparison between the cumulative uncertainty residual for a team of UAVs controlled by our method and a team of UAVs controlled using the coverage and tracking controller framework presented by \cite{pham2017distributed}. For cases 1, 2, and 3, our approach was able to achieve significantly lower cumulative uncertainty.






\section{Conclusion}
\label{sec:discussion}

We have introduced a novel analytical bound on fire propagation uncertainty, allowing high-quality path planning for real-time wildfire monitoring and tracking, while also providing a probabilistic guarantee on human safety. Our approach outperformed prior work for distributed control of UAVs for wildfire tracking, as well as a reinforcement learning baseline. In future work, we aim to relax the simplifying assumptions of the FARSITE model and to further explore learning-based techniques.

\bibliographystyle{named}
\bibliography{paper}


\section*{Appendix}
\label{sec:appendix}
\subsubsection{Deriving the Upper-bound Time for Wildfire Case 3 (e.g., $ T_{UB}^{(C3)} $)}
\label{subsubsec:appndx_T_UB3}
\begin{figure}[t!]
	\centering
	\includegraphics[trim=0in 0in 0in 0in,clip,width=0.6\columnwidth]{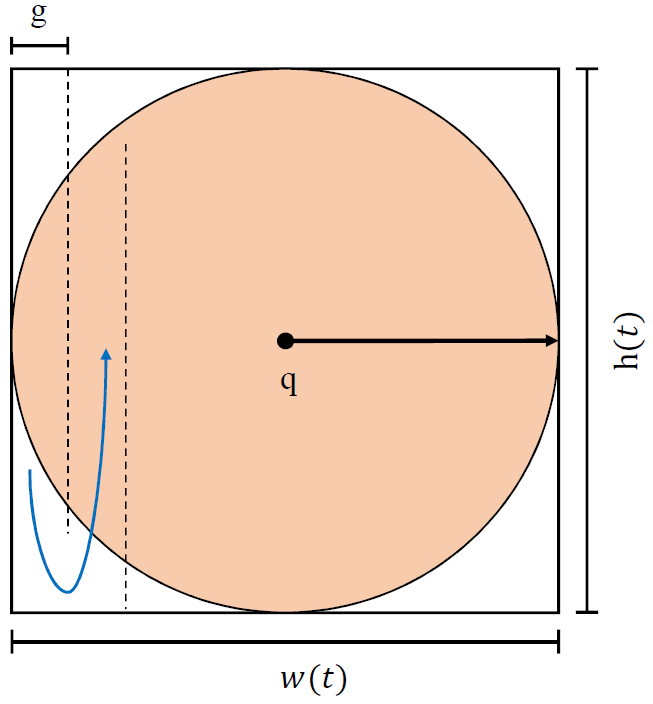}
	\caption{The third wildfire scenario, in which a fire can move and spawn quickly. The large red circle shows the potential area engulfed by the growth of each firefront. Parameter $ g $ represents the width of the UAV's FOV. The blue arrow represents a UAV's partial trajectory while monitoring a fire (vertical scanning pattern is assumed).}
	\label{fig:appndx1}
\end{figure}

To arrive at this bound, consider Figure~\ref{fig:appndx1} as the case detailed in Section~\ref{subsubsec:rapidly-growing-fire}. According to Figure~\ref{fig:appndx1}, for a specific fire, $q$ the time-varying width and height (i.e., planar length) of the enlarging fire area along \textit{X} and \textit{Y} axes can be estimated as in Equations~\ref{eq:w_t} and~\ref{eq:h_t} at the $\alpha$ confidence level, as fires are now allowed to spread for $T_{UB}^{(\text{C3})}$ units of time
\par\nobreak{\parskip0pt \footnotesize \noindent
	\begin{align}
		w(t) \leq & ~2\left.\reallywidehat{\dot{q}_t^x}\right|_{\alpha}T_{UB}^{(C3)} \label{eq:w_t}\\
		h(t) \leq & ~2\left.\reallywidehat{\dot{q}_t^y}\right|_{\alpha}T_{UB}^{(C3)} \label{eq:h_t}
\end{align}} Assuming a vertical scanning pattern as shown, we round up the maximum possible $ w(t) $ and thus, the total number of passes the drone would take from left to right is given by
\par\nobreak{\parskip0pt \footnotesize \noindent
	\begin{align}
		n(t) \leq \ceil[\Bigg]{ \frac{2\left.\reallywidehat{\dot{q}_t^x}\right|_{\alpha}T_{UB}^{(C3)}}{g}} \label{eq:n_t}
\end{align}}and the total distance traversed for each pass is
\par\nobreak{\parskip0pt \footnotesize \noindent
	\begin{align}
		d(t) = 2\left.\reallywidehat{\dot{q}_t^y}\right|_{\alpha}T_{UB}^{(C3)} \label{eq:d_t}
\end{align}}and finally the total pass traversed is
\par\nobreak{\parskip0pt \footnotesize \noindent
	\begin{align}
		d^{tot}(t) = n(t)h(t) \label{eq:d_total}
\end{align}}Accordingly, the time required for one firefront point to escape the FOV of a specific drone $ d $ with velocity $ v $ can be calculated as
\par\nobreak{\parskip0pt \footnotesize \noindent
	\begin{align}
		\tau_q(t) = & \frac{d^{tot}(t)}{v} \label{eq:time_required}\\
		= & \frac{n(t)h(t)}{v}\\
		= & \frac{\ceil[\Bigg]{ \frac{2\left.\reallywidehat{\dot{q}_t^x}\right|_{\alpha}T_{UB}^{(C3)}}{g}}2\left.\reallywidehat{\dot{q}_t^y}\right|_{\alpha}T_{UB}^{(C3)}}{v} \label{eq:time_required_final}
\end{align}}Now, the time required for all of the fire points will be the summation over all $ \tau_q $ in Equation~\ref{eq:time_required_final}. However, we must note that the center of each spreading fire point also moves, and the upper bound time derived for case 2 in Equation \ref{eq:T_UBcase2} (i.e., $ T_{UB}^{(\text{C2})} $) must be added to this summation. Therefore, the upper bound time for case 3 can be estimated as in Equation~\ref{eq:T_UBCase3_start_appx}.
\par\nobreak{\parskip0pt \footnotesize \noindent
	\begin{align}
		T_{UB}^{(C3)} &=  \left.T_{UB}^{(\text{C2})}\right|_{C3} +  \sum_q \frac{2}{v}\left.\reallywidehat{\dot{q}_t^y}\right|_{\alpha}T_{UB} \ceil[\Bigg]{ \frac{2\left.\reallywidehat{\dot{q}_t^x}\right|_{\alpha}T_{UB}^{(C3)}}{g}} \label{eq:T_UBCase3_start_appx}
\end{align}}We need to solve Equation~\ref{eq:T_UBCase3_start_appx} for $ T_{UB}^{(C3)} $ to find the final equation to obtain the upper bound. For this purpose, we make the following two simplifying assumptions. First, we remove the ceiling operator and add one to the term inside the operator to achieve continuity
, as shown in Equation \ref{eq:ceil}. 
\par\nobreak{\parskip0pt \footnotesize \noindent
	\begin{align}
		\ceil[\Bigg]{ \frac{2\left.\reallywidehat{\dot{q}_t^x}\right|_{\alpha}T_{UB}^{(C3)}}{g}}\leq { \frac{2\left.\reallywidehat{\dot{q}_t^x}\right|_{\alpha}T_{UB}^{(C3)}}{g}} + 1 \label{eq:ceil}
\end{align}}
Second, we adopt a similar approach to case 2 by assuming the area required to cover to account for the growth of each fire location is upper-bounded by $|Q|$ times the area of growth for a hypothetical fire growing quickest
(Equation \ref{eq:fastest}).
\par\nobreak{\parskip0pt \footnotesize \noindent
	\begin{align}
		&\sum_q \left.\reallywidehat{\dot{q}_t^y}\right|_{\alpha}T_{UB} \ceil[\Bigg]{ \frac{2\left.\reallywidehat{\dot{q}_t^x}\right|_{\alpha}T_{UB}^{(C3)}}{g}} \leq  |Q|\zeta^{\alpha} \ceil[\Bigg]{ \frac{2\zeta^{\alpha}T_{UB}^{(C3)}}{g}} \label{eq:fastest}
\end{align}} 
With these two conservative simplifying assumptions, the bound in Equation \ref{eq:T_UBCase3_start_appx} can be revised as below.
\par\nobreak{\parskip0pt \footnotesize \noindent
\begin{align}
		T_{UB}^{(C3)} &=  \left.T_{UB}^{(\text{C2})}\right|_{C3} +  \frac{2|Q|\zeta^{\alpha}T_{UB}^{(C3)}}{v}\left[ \frac{2\zeta^{\alpha}T_{UB}^{(C3)}}{g} +1 \right]  \label{eq:T_UBCase3_start_appx_revised} \\ 
		T_{UB}^{(C3)} &- \frac{2|Q|\zeta^{\alpha}T_{UB}^{(C3)}}{v}\left[ \frac{2\zeta^{\alpha}T_{UB}^{(C3)}}{g} +1 \right] = \left.T_{UB}^{(\text{C2})}\right|_{C3} \\ 
		T_{UB}^{(C3)} &- \frac{2|Q|\zeta^{\alpha}T_{UB}^{(C3)}}{v}\left[ \frac{2\zeta^{\alpha}T_{UB}^{(C3)}}{g} +1 \right] = \frac{MST}{\frac{v}{2} - 2\zeta^{\alpha}(|Q|-1)} \label{eq:T_UBCase3_start_appx_revised1}
\end{align}}
Note that Equation~\ref{eq:T_UBCase3_start_appx_revised1} is in the form of $ z-az\left(bz+1\right) =\delta $ general quadratic equations with $ z= T_{UB}^{(C3)} $, which can be simplified to $ \gamma z^2 - \beta z + \delta = 0 $, where $ \gamma = ab $ and $ \beta=1+a $ and they can be calculated as below.
\par\nobreak{\parskip0pt \footnotesize \noindent
	\begin{align}
		\gamma &=  \frac{4|Q|\left(\zeta^{\alpha}\right)^2}{gv}\\
		\beta &= 1 + \frac{2|Q|\zeta^{\alpha}}{v}      \\
		\delta &= \frac{MST}{\frac{v}{2} - 2\zeta^{\alpha}(|Q|-1)}
\end{align}}Eventually, from the solution to general quadratic equations, we know that the upper-bound traversal time $ T_{UB}^{(C3)} $ allowed to maintain the track quality of each fire for case 3 can be obtained as in Equation~\ref{eq:T_UBCase3_app}.
\par\nobreak{\parskip0pt \footnotesize \noindent
	\begin{align}
		T_{UB}^{(C3)} = \frac{-\beta + \sqrt{\beta^2 - 4\gamma\delta}}{2\gamma} \label{eq:T_UBCase3_app}
\end{align}}

\subsubsection{Calculating the Jacobian Matrices and the Gradients}
\label{subsubsec:appndx_jacobians}
To calculate, we first reform the state transition equation in~\ref{eq:probform1} to account for all aforementioned associating parameters as below.
\par\nobreak{\parskip0pt \footnotesize \noindent
	\begin{equation}
		\label{eq:probforrm2}
		\Bigg[ \mathcal{S}_t \Bigg]_{8\times1} =
		\Bigg[ \left. \frac{\partial f}{\partial \mathcal{S}_i}\right|_{\hat{\mathcal{S}}_{t-1|t-1}} \Bigg]_{8\times8}
		\Bigg[ \mathcal{S}_{t-1} \Bigg]_{8\times1} + \omega_t
\end{equation}}where the process noise $  \omega_t $ should account for both stochasticity in fire behavior and propagation model inaccuracy and is modeled by a zero mean white Gaussian noise. Accordingly, we form the state transition Jacobian matrices $ F_t $ in Equation \ref{eq:statetransitionjacobian}, including partial derivatives of wildfire propagation dynamics in Equations~\ref{eq:firedynamics1} and~\ref{eq:qdotdefinition}, with respect to all variables in state vector $ \mathcal{S}_t $. Note that the parameters $ R_t $, $ U_t $, and $ U_t $ are not necessarily dynamic with time, and it is fairly reasonable to consider these physical parameters as constants for short periods of time. However, in the case of analyzing the system for longer durations, temporal dynamics may apply (\cite{delamatar2013downloading}), specifically due to changes in wind speed and velocity. Exact estimation of temporal dynamics related to these parameters are out of the scope of the current study, since we assume locality in time and space according to FARSITE~(\cite{finney1998farsite}). The resulting Jacobian matrix will take the form as in Equation~\ref{eq:statetransitionJacob}. To calculate the partial derivatives in Equation~\ref{eq:statetransitionJacob}, we need the fire dynamics through wildfire mathematical model. By ignoring the superscript $ i $ in Equations~\ref{eq:firedynamics1} and~\ref{eq:qdotdefinition} and without losing generality, $ \dot{q}_{t} $ can be estimated for each propagating firefront by Equations \ref{eq:qdot1}-\ref{eq:c}, where $ \dot{q}_{t}^x $ and $ \dot{q}_{t}^x $ are firefront first-order dynamic for \textit{X} and \textit{Y} coordinates, and $ a $, $ b $, $ c $, $ d $, and $ l $ are constants
\par\nobreak{\parskip0pt \footnotesize \noindent
	\begin{align}
		\dot{q}_{t}^x &= C(R_t, U_t)\sin(\theta_t) \label{eq:qdot1} \\
		\dot{q}_{t}^y &= C(R_t, U_t)\cos(\theta_t)     \label{eq:qdot2}
\end{align}} where
\par\nobreak{\parskip0pt \footnotesize \noindent
	\begin{align}
		C(R_t, U_t) &= R_t\left(1-\frac{LB(U_t)}{LB(U_t) + \sqrt{GB(U_t)}}\right) \label{eq:c}
\end{align}} and
\par\nobreak{\parskip0pt \footnotesize \noindent
	\begin{align}
		LB(U_t) &= ae^{bU_t} + ce^{-dU_t} + l \label{eq:lb} \\
		GB(U_t) &= LB(U_t)^2 - 1 \label{eq:gb}
\end{align}}
\noindent Now, the derivatives of $ q_t^x $ and $ q_t^y $ with respect to parameters $ R_{t-1} $, $ U_{t-1} $, and $ \theta_{t-1} $ are computed by applying the chain-rule and using Equation~\ref{eq:qdot1}-\ref{eq:gb} as follows, where $ \mathcal{D}(\theta) $ equals $ \sin\theta $ and $ \cos\theta $ for X and Y axis, respectively.
\par\nobreak{\parskip0pt \footnotesize \noindent
	\begin{align}
		\frac{\partial q_{t}}{\partial \theta_{t-1}} &= ~C(R_t, U_t)\frac{\partial\mathcal{D}(\theta)}{\partial\theta}\delta t \label{eq:qJacobs1}\\
		\frac{\partial q_{t}}{\partial R_{t-1}} &= \left(1-\frac{LB(U_t)}{LB(U_t) + \sqrt{GB(U_t)}}\right)\mathcal{D}(\theta)\delta t \\
		\frac{\partial q_{t}}{\partial U_{t-1}} &= \frac{R_{t'}\bigg(LB(U_{t'})\frac{\partial GB(U_{t'})}{\partial U_{t'}} - GB(U_{t'})\frac{\partial LB(U_{t'})}{\partial U_{t'}}\bigg)}{\left(LB(U_{t'})+\sqrt{GB(U_{t'})}\right)^2}\mathcal{D}(\theta)\delta t \label{eq:qJacobs2}
\end{align}} 

\begin{figure}[t!]
	\centering
	\includegraphics[trim=0in 0in 0in 0in,clip,width=0.8\columnwidth]{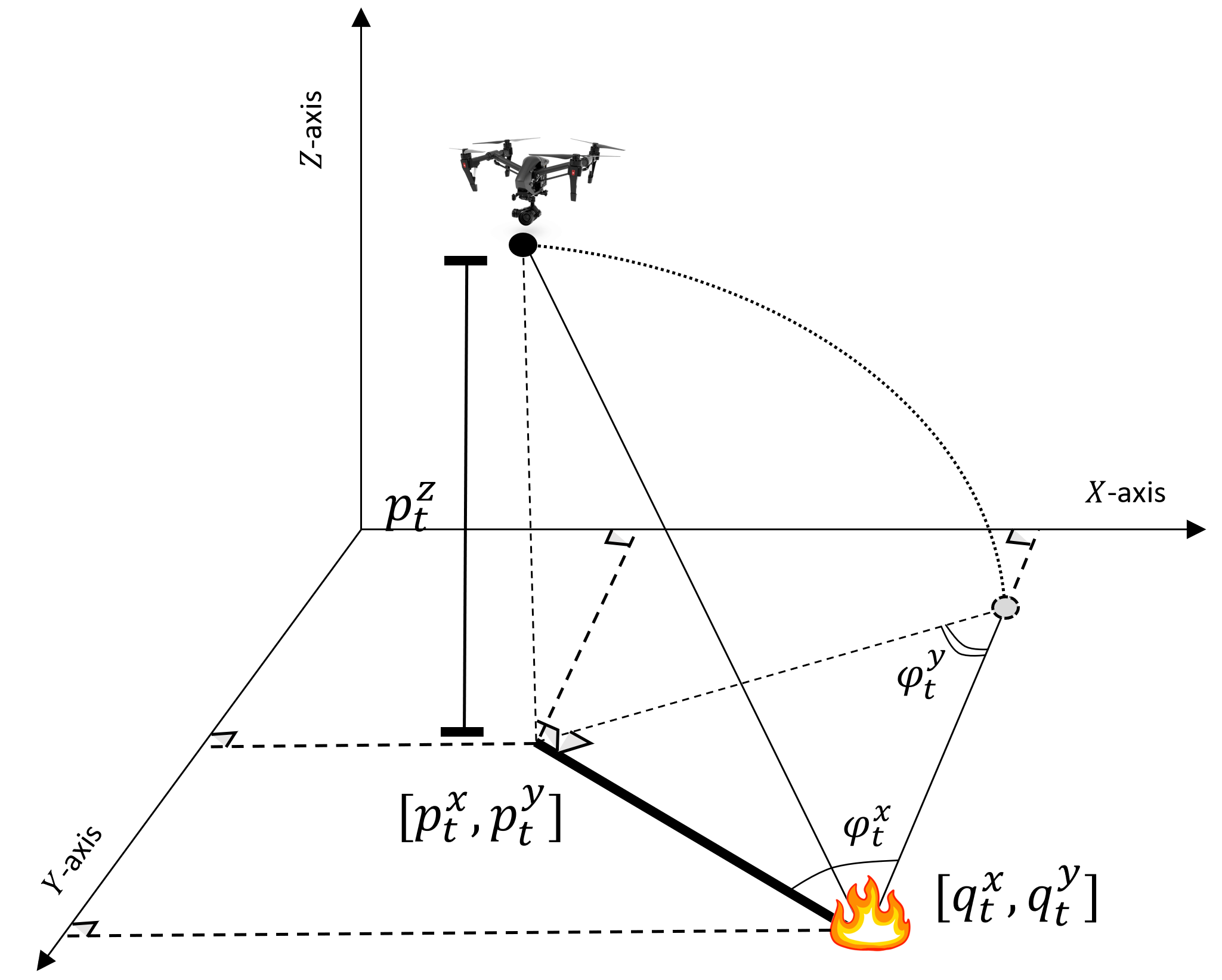}
	\caption{The UAV observation model.}
	\label{fig:ObsMdl}
\end{figure}

\noindent To calculate the observation Jacobian matrix $ H_t $ in Equation~\ref{eq:observationjacob} and its gradients, we first consider Figure~\ref{fig:ObsMdl} which depicts the observation model through which UAVs perceive afire.

Accordingly, the observation mapping equation in~\ref{eq:probform3} is now presented as
\par\nobreak{\parskip0pt \footnotesize \noindent
	\begin{equation}
		\label{eq:probforrm3}
		\Bigg[ \hat{\Phi}_t \Bigg]_{5\times1}
		=
		\Bigg[ \left. \frac{\partial h}{\partial \Phi_i}\right|_{\Phi_{t|t}} \Bigg]_{5\times8}
		\Bigg[ \Phi_{t} \Bigg]_{8\times1} + \nu_t
\end{equation}}The angle parameters (i.e., $ \varphi_t^x $ and $ \varphi_t^y $) contain information regarding both firefront location $ [q_t^x, q_t^y] $ and UAV coordinates $ [p_t^x, p_t^y] $. Considering $ p_t^d = \left[p_{t}^x, p_{t}^y, p_{t}^z\right]^T $ as the desired current UAV pose, there will be errors associated with all of these coordinates. This will affect UAVs’ ability to extrapolate where afire is on the ground. Taking this into account is very important. The observation noise is modeled as a zero-mean white Gaussian noise. Note that errors in X, Y, and Z axes coordinates of a drone are loosely correlated, and thus, we also incorporate non-diagonal elements in observation noise covariance matrix $ R_t $. According to Figure~\ref{fig:ObsMdl}, by projecting the looking vector of a UAV to planar coordinates, the angle parameters are calculated as below
\par\nobreak{\parskip0pt \footnotesize \noindent
	\begin{align}
		\label{eq:angleparamx}
		\varphi_t^x = & \tan^{-1}\left(\frac{q_t^x - p_t^x}{p_t^z}\right) \\
		\label{eq:angleparamy}
		\varphi_t^y = & \tan^{-1}\left(\frac{q_t^y - p_t^y}{p_t^z}\right)
\end{align}}. The observation Jacobian matrix $ H_t $ is presented in Equation~\ref{eq:observationjacob}, where the partial derivatives are derived using Equations~\ref{eq:angleparamx} and~\ref{eq:angleparamy} as follows
\par\nobreak{\parskip0pt \footnotesize \noindent
	\begin{align}
		\label{eq:Hjacobsxaa}
		\frac{\partial \varphi_{t}^x}{\partial q_t^x} = & \frac{1}{1+\left(\frac{q_t^x-p_t^x}{p_t^z}\right)^2}\left(\frac{1}{p_t^z}\right)\\
		\frac{\partial \varphi_{t}^x}{\partial p_t^x} = & \frac{1}{1+\left(\frac{q_t^x-p_t^x}{p_t^z}\right)^2}\left(\frac{-1}{p_t^z}\right)\\
		\frac{\partial \varphi_{t}^x}{\partial p_t^z} = & \frac{1}{1+\left(\frac{q_t^x-p_t^x}{p_t^z}\right)^2}\left(q_t^x-p_t^z\right)\left(\frac{-1}{(p_t^z)^2}\right) \label{eq:Hjacobsxx}
\end{align}}similar equations as above hold for \textit{Y}-axis with $ q_t^y $ and $ p_t^y $.





\end{document}